\documentclass[twocolumn]{article}

\usepackage[width=17cm,height=22cm]{geometry}
\usepackage[T1]{fontenc}
\usepackage[utf8]{inputenc}
\usepackage{fancyvrb}
\usepackage{booktabs}
\usepackage{authblk}
\usepackage{url}
\usepackage{graphicx}
\usepackage{xcolor}
\usepackage{tikz}
\usetikzlibrary{matrix,calc}
\usetikzlibrary{decorations.pathreplacing}

\tikzstyle{overbrace text style}=[font=\tiny, above, pos=.5, yshift=5mm]
\tikzstyle{overbrace style}=[decorate,decoration={brace,raise=2mm,amplitude=3pt}]
\tikzstyle{underbrace style}=[decorate,decoration={brace,raise=2mm,amplitude=3pt,mirror},color=gray]
\tikzstyle{underbrace text style}=[font=\tiny, below, pos=.5, yshift=-5mm]

\definecolor{lightblue}{rgb}{0.145,0.6666,1}
\definecolor{magenta}{RGB}{5,122,255}
\definecolor{frenchrose}{RGB}{255,71,133}
\definecolor{phlox}{RGB}{230,8,255}
\definecolor{carmine}{RGB}{181,63,63}

\title{CAp 2017 challenge: Twitter Named Entity Recognition }

\author[1]{Cédric Lopez\thanks{For any details you may contact: cedric.lopez@viseo.com, ioannis.partalas@gmail.com}}
\author[2]{Ioannis Partalas}
\author[3]{Georgios Balikas}
\author[1]{Nadia Derbas}
\author[4]{Amélie Martin}
\author[4]{Coralie Reutenauer}
\author[1]{Frédérique Segond}
\author[3]{Massih-Reza Amini}

\affil[1]{Viseo R\&D, 4, avenue doyen Louis Weil, Grenoble, France}
\affil[2]{Expedia, Inc}
\affil[3]{Université Grenoble-Alpes}
\affil[4]{SNCF}

\begin{document}
\maketitle

\begin{abstract}
The paper describes the CAp 2017 challenge. 
The challenge concerns the problem of Named Entity Recognition (NER) for tweets written in French. 
We first present the data preparation steps we followed for constructing the dataset released in the framework of the challenge. We begin by demonstrating why NER for tweets is a challenging problem especially when the number of entities increases. We detail the annotation process and the necessary decisions we made. We provide statistics on the inter-annotator agreement, and we conclude the data description part with examples and statistics for the data. We, then, describe the participation in the challenge, where 8 teams participated, with a focus on the methods employed by the challenge participants and the scores achieved in terms of F$_1$ measure. Importantly, the constructed dataset comprising $\sim$6,000 tweets  annotated for 13 types of entities, which to the best of our knowledge is the first such dataset in French, is publicly available at \url{http://cap2017.imag.fr/competition.html} .
\end{abstract}

\medskip


\section{Introduction}
\label{sec:intro}
The proliferation of the online social media has lately resulted in the democratization of online content sharing. Among other media, Twitter is very popular for research and application purposes due to its scale, representativeness and ease of public access to its content. However, tweets, that are short messages of up to 140 characters, pose several challenges to traditional Natural Language Processing (NLP) systems due to the creative use of characters and punctuation symbols, abbreviations ans slung language.

Named Entity Recognition (NER) is a fundamental step for most of the information extraction pipelines. Importantly, the terse and difficult text style of tweets presents serious challenges to NER systems, which are usually trained using more formal text sources such as newswire articles or Wikipedia entries that follow particular morpho-syntactic rules. As a result, off-the-self tools trained on such data perform poorly \cite{ritter2011named}. The problem becomes more intense as the number  of entities to be identified increases, moving from the traditional setting of very few entities (persons, organization, time, location) to problems with more. Furthermore, most of the resources (\textit{e.g.}, software tools) and benchmarks for NER are for text written in English. As the multilingual content online increases,\footnote{\url{http://www.internetlivestats.com/internet-users/#byregion}} and English may not be anymore the \textit{lingua franca} of the Web. Therefore,  having resources and benchmarks in other languages is crucial for enabling information access worldwide.   

In this paper, we propose a new benchmark for the problem of NER for tweets written in French. The tweets were collected using the publicly available Twitter API and annotated with 13 types of entities. The annotators were native speakers of French and had previous experience in the task of NER. Overall, the generated datasets consists of $\sim 6,000$ tweets, split in training and test parts.

The paper is organized in two parts. In the first, we discuss the data preparation steps (collection, annotation) and we describe the proposed dataset. The dataset was first released in the framework of the CAp 2017 challenge, where 8 systems participated. Following, the second part of the paper presents an overview of baseline systems and the approaches employed by the systems that participated. We conclude with a discussion of the performance of Twitter NER systems and remarks for future work.

% \section{Related Works}
% \label{sec:relworks}

\section{Challenge Description}
\label{sec:challenge Description}
In this section we describe the steps taken during the organisation of the challenge. We begin by introducing the general guidelines for participation and then proceed to the description of the dataset. 

\subsection{Guidelines for Participation}

The CAp 2017 challenge concerns the problem of NER for tweets written in French. A significant milestone while organizing the challenge was the creation of a suitable benchmark. 
While one may be able to find Twitter datasets for NER in English, to the best of our knowledge, this is the first resource for Twitter NER in French. 
Following this observation, our expectations for developing the novel benchmark are twofold: first, we hope that it will further stimulate the research efforts for French NER with a focus on in user-generated text social media. 
Second, as its size is comparable with datasets previously released for English NER we expect it to become a reference dataset for the community.

The task of NER decouples as follows: given a text span like a tweet, one needs to identify contiguous words within the span that correspond to entities. Given, for instance, a tweet ``Les Parisiens supportent PSG ;-)'' one needs to identify that the abbreviation ``PSG'' refers to an entity, namely the football team ``Paris Saint-Germain''. Therefore, there two main challenges in the problem. First one needs to identify the boundaries of an entity (in the example PSG is a single word entity), and then to predict the type of the entity. 
In the CAp 2017 challenge one needs to identify among 13 types of entities: person, musicartist, organisation, geoloc, product, transportLine, media, sportsteam, event, tvshow, movie, facility, other in a given tweets. Importantly, we do not allow the entities to be hierarchical, that is contiguous words belong to an entity as a whole and a single entity type is associated per word.
It is also to be noted that some of the tweets may not contain entities and therefore systems should not be biased towards predicting one or more entities for each tweet. 

Lastly, in order to enable participants from various research domains to participate, we allowed the use of any external data or resources. On one hand, this choice would enable the participation of teams who would develop systems using the provided data or teams with previously developed systems capable of setting the state-of-the-art performance.  On the other hand, our goal was to motivate approaches that would apply transfer learning or domain adaptation techniques on already existing systems to adapt them for the task of NER for French tweets.

\subsection{The Released Dataset}
\label{sec:dataset}

For the purposes of the CAp 2017 challenge we constructed a dataset for NER of French tweets. Overall, the dataset comprises 6,685 annotated tweets with the 13 types of entities presented in the previous section. The data were released in two parts: first, a training part was released for development purposes (dubbed ``Training'' hereafter). Then, to evaluate the performance of the developed  systems a ``Test'' dataset was released that consists of 3,685 tweets. For compatibility with previous research, the data were released tokenized using the CoNLL format and the BIO encoding.

To collect the tweets that were used to construct the dataset we relied on the Twitter streaming  API.\footnote{\url{https://dev.twitter.com/rest/public}} The API makes available a part of Twitter flow and one may use particular keywords to filter the results. In order to collect tweets written in French and obtain a sample that would be unbiased towards particular types of entities we used common French words like articles, pronouns, and prepositions: ``le'',``la'',``de'',``il'',``elle'', etc.. In total, we collected 10,000 unique tweets from September 1st until September the 15th of 2016. 

Complementary to the collection of tweets using the Twitter API, we used 886 tweets  provided by the ``Société Nationale des Chemins de fer Français'' (SNCF), that is the French National Railway Corporation. The latter subset is biased towards information in the interest of the corporation such as train lines or names of train stations. To account for the different distribution of entities in the tweets collected by SNCF we  incorporated them in the  data as follows: 
\begin{itemize}
	\item For the training set, which comprises 3,000 tweets, we used 2,557 tweets collected using the API and 443 tweets of those provided by SNCF.
	\item For the test set, which comprises 3,685 consists we used 3,242 tweets from those collected using the API and the remaining 443 tweets from those provided by SNCF.
\end{itemize}

\section{Annotation}
\label{sec:annotation}
In the framework of the challenge, we were required to first identify the entities occurring in the dataset and, then, annotate them with of the 13 possible types. Table \ref{tbl:entities_descr} provides a description for each type of entity that we made available both to the annotators and to the participants of the challenge.  

Mentions  (strings starting with @) and hashtags (strings starting with \#) have a particular function in tweets. The former is used to refer to persons while the latter to indicate keywords. Therefore, in the annotation process we treated them using the following protocol: A hashtag or a mention should be annotated as an entity if:
\begin{enumerate}
 \item the token is an entity of interest, and 
 \item the token has a syntactic role in the message
\end{enumerate}
For a hashtag or a mention to be annotated both conditions are to be met. Figure \ref{ex2} elaborates on that:
\begin{itemize}
	\item ``@\_x\_Cecile\_x'' is not annotated because this is expressly a mention referring to a twitter account: it does not play any role in the tweet (except to indicate that it is a retweet of that person).
	\item ``cécile'' is annotated because this is a named entity of type \textit{person}
	\item ``\#cecile'' is a person but it does not play a syntactic role: it plays here the role of metadata marker.
\end{itemize}

% Figure \ref{fig:entity_distributions} shows the distribution of entities in the training and test parts of the dataset. Notice that the distribution of entities across the entity types is very similar between the training and the test data. The most common entity type is ``Person'' followed by ``Geoloc'' and ``Transport Line''. As ``Transport Line'' was an entity with particular interest for SNCF who provided part of the labeled data, we expect it to occur less frequently in a random sample of tweets. This does not impact our evaluation process however as the figure clearly demonstrates that the distributions of training and test data are similar. [Balikas: I am here.]

\begin{table*} \scriptsize
 \begin{tabular}{l l}
 \toprule
 Entity Type & Description \\
 \midrule
  Person & \multicolumn{1}{p{14cm}}{First name, surname, nickname, film character. Examples: \textit{JL Debré, Giroud, Voldemort}. The entity type does not include imaginary persons, that should be instead annotated as ``Other'' }\\[.4cm]
  Music Artist & \multicolumn{1}{p{14cm}}{Name and (or) surname of music artists. Examples: \textit{Justin Bieber}. The category also includes names of music groups, musicians and composers.} \\[.4cm]
  Organization & \multicolumn{1}{p{14cm}}{Name of organizations. Examples: \textit{Pasteur}, \textit{Gien}, \textit{CPI}, \textit{Apple}, \textit{Twitter}, \textit{SNCF}}\\[.4cm]
  
Geoloc &\multicolumn{1}{p{14cm}}{Location identifiers. Examples: \textit{Gaule}, \textit{mont de Tallac}, \textit{France}, \textit{NYC} }\\[.4cm]

Product & \multicolumn{1}{p{14cm}}{Product and brand names. A product is the result of human effort including video games, titles of songs,  titles of  Youtube videos, food, cosmetics, etc., Movies are excluded (see \textit{movie} category). Examples: \textit{Fifa17}, \textit{Canapé convertible 3 places SOFLIT}, \textit{Cacharel}, \textit{Je meurs d’envie de vous} (Daniel Levi's song).}\\[.4cm]

Transport Line & \multicolumn{1}{p{14cm}}{Transport lines. Examples: \textit{@RERC\_SNCF}, \textit{ligne j}, \textit{\#rerA}, \textit{bus 21}}\\[.4cm]

Media & \multicolumn{1}{p{14cm}}{All media aiming at disseminating information.  The category also includes organizations that publish user content. Examples:\textit{Facebook}, \textit{Twitter}, \textit{Actu des People}, \textit{20 Minutes}, \textit{AFP}, \textit{TF1}}\\[.4cm]

Sportsteam &  \multicolumn{1}{p{14cm}}{Names of sports teams. Example: \textit{PSG}, \textit{Troyes}, \textit{Les Bleus}.}\\[.4cm]

Event & \multicolumn{1}{p{14cm}}{Events. Examples:  \textit{Forum des Entreprises ILIS}, \textit{Emmys}, \textit{CAN 2016}, \textit{Champions League}, \textit{EURO U17}}\\[.4cm]

Tvshow & \multicolumn{1}{p{14cm}}{Television emission like morning shows, reality shows and television games. It excludes films and tvseries (see \textit{movie} category). Examples: Secret Story, Motus, Fort Boyard, Télématin }\\[.4cm]

Movie & \multicolumn{1}{p{14cm}}{Movies and TV series. Examples:  \textit{Game of Thrones}, \textit{Jumanji}, \textit{Avatar}, \textit{La traversée de Paris} }\\[.4cm]

Facility & \multicolumn{1}{p{14cm}}{Facility. Examples:  \textit{OPH de Bobigny}, \textit{Musée Edmond Michelet}, \textit{Cathédrale Notre-Dame de Paris}, \textit{@RER\_C}, \textit{la ligne C}, \textit{\#RERC}. Train stations are also part of this category, for example \textit{Evry Courcouronnes}. }\\[.4cm]

Other & \multicolumn{1}{p{14cm}}{Named entities that do not belong in the previous categories. Examples:  \textit{la Constitution}, \textit{Dieu}, les \textit{Gaulois} }\\[.4cm]
  \bottomrule
 \end{tabular}
\caption{Description of the 13 entities used to annotate the tweets. The description was made available to the participants for the development of their systems. }\label{tbl:entities_descr}
\end{table*}

\begin{figure*}\centering
   \includegraphics[scale=0.6]{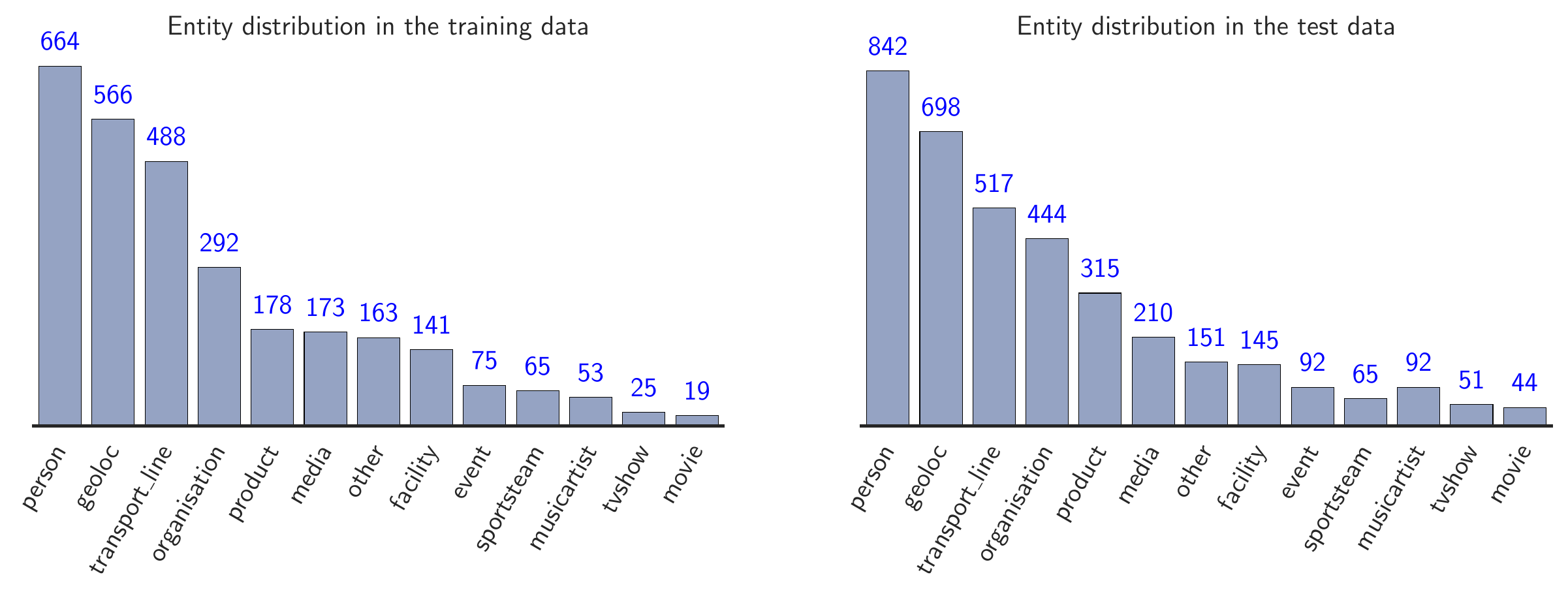}
      \caption{ Distribution of entities across the 13 possible entity types for the training and test data. Overall,  2,902 entities occur in the training data and 3,660 in the test.}\label{fig:entity_distributions}
\end{figure*}

% All in all, 6,685 tweets were manually annotated by 4 organizers of the CAP competition who used the BIO encoding (Begin-In-Out) (cf. \ref{ex1}. 

\begin{figure}
   \caption{\label{ex1} Example of an annotated tweet.}
   \includegraphics{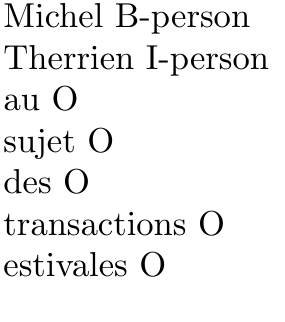}
\end{figure}

We measure the inter-annotator agreement between the annotators based on the Cohen's Kappa (cf. Table \ref{tbl:cohens}) calculated on the first 200 tweets of the training set. According to \cite{landis1977application} our score for Cohen's Kappa (0,70) indicates a strong agreement.

\begin{table}[t]\small
    \centering
    \begin{tabular}{l cccc }
    \toprule
    & Ann$_1$ & Ann$_2$ & Ann$_3$ & Ann$_4$ \\\midrule
    Ann$_1$ & - & 0.79& 0.71& 0.63   \\
    Ann$_2$ &0.79 &-& 0.64& 0.61 \\
    Ann$_3$ &0.71&0.64&-& 0.55\\
    Ann$_4$ &0.63&0.61&0.55&- \\
    \bottomrule
    \end{tabular}
    \caption{Cohen's Kappa for the interannotator agreement. ``Ann" stands for the annotator. The Table is symmetric.}
    \label{tbl:cohens}
\end{table}

% Table \ref{tab:nbentity} shows the number of annotated entities for each category. 6,562 entities were ennotated.

% \begin{table}
% 	\centering
% 		\begin{tabular}{|l|c|c|c|}
%   \hline
%   Categories & Training set & Test set & Total \\
%   \hline
%   person & 664 & 842 & 1506 \\
%   musicartist & 53 & 92 & 145 \\
% 	organisation & 292 & 444 & 736 \\
% 	geoloc & 566 & 698 & 1264 \\
% 	product & 178 & 315 & 493 \\
% 	media & 173 & 210 & 383 \\
% 	sportsteam & 65 & 65 & 130 \\
% 	event & 75 & 92 & 167 \\
% 	tvshow & 25 & 51 & 76 \\
% 	movie & 19 & 44 & 63 \\
% 	transportLine & 488 & 517 & 1005 \\
% 	facility & 141 & 145 & 286 \\
% 	other & 163 & 151 & 314 \\
% 	\textbf{Total} & \textbf{2902} & \textbf{3660} & \textbf{6562} \\
%   \hline
% \end{tabular}
% 	\caption{Number of entities for each category.}
% 	\label{tab:nbentity}
% \end{table}

\begin{figure}
   \caption{\label{ex2} Exemple of an annotated tweet.}
   \includegraphics{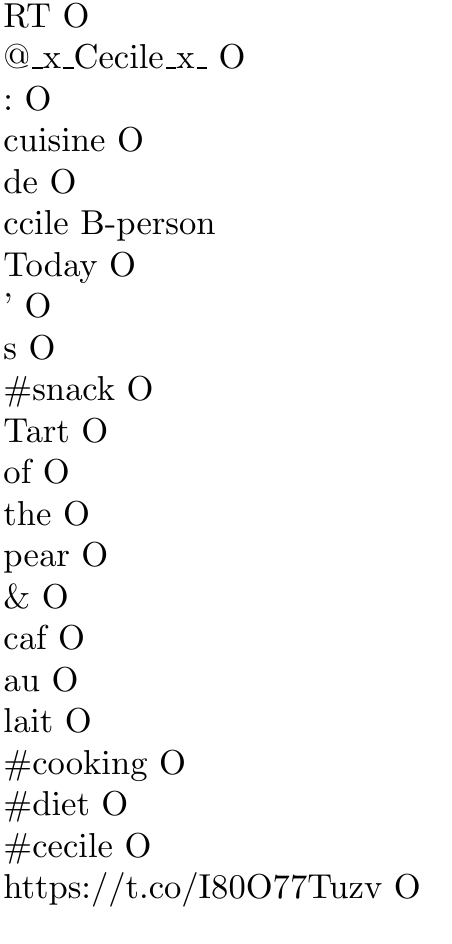}
\end{figure}

\begin{figure}
   \caption{\label{ex3} Exemple of an annotated tweet.}
   \includegraphics{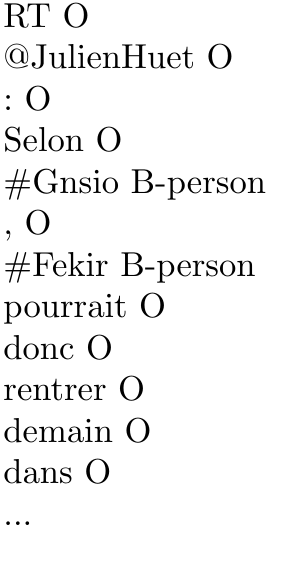}
\end{figure}

In the example given in Figure \ref{ex3}:
\begin{itemize}
	\item \textit{@JulienHuet} is not annotated because this is a reference to a twitter account without any syntactic role. 
	\item \textit{\#Génésio} is annotated because it is a hashtag which has a syntactic role in the message, and it is of type \textit{person}.
	\item \textit{Fekir} is annotated following the same rule that \textit{\#Génésio}.
\end{itemize}

 \begin{tikzpicture}
     \matrix[name=M1, matrix of nodes, row sep=10pt, column sep=3pt,ampersand replacement=\&]{
 	    \node (schema) [text=black] {Il}; \& \node (schema-spezifisch) [text=black] {rejoint}; \& \node (nutzerinfo) [text=frenchrose] {Pierre}; \& \node (host) [text=frenchrose] {Fabre}; \\  \node (query) [text=black] {comme}; \& \node (fragment) [text=black] {directeur};
 	    \& \node [text=black] {des}; \& \node [text=black] {marques};\\  \node(ducray) [text=magenta] {Ducray}; \& \node [text=black] {et}; \& \node(a) [text=magenta] {A};\& \node [text=magenta] {-}; \&
 	    \node(derma) [text=magenta] {Derma};\\
 		        };
 			\draw [overbrace style] (nutzerinfo.north west) -- (host.north east) node [overbrace text style,rectangle,draw,color=white,rounded corners,inner sep=4pt,
 			fill=frenchrose] {\Large\textbf{Group}};
 			\draw [underbrace style] (ducray.south west) -- (ducray.south east) node [underbrace text style,rectangle,draw=black,color=white,rounded corners,inner sep=4pt,
 						fill=magenta] {\Large\textbf{Brand}};
 			\draw [underbrace style] (a.south west) -- (derma.south east) node [underbrace text style,rectangle,draw,color=white,rounded corners,inner sep=4pt,
 						fill=magenta] {\Large\textbf{Brand}};
 %			\draw[thick,dotted]     ($(nutzerinfo.north west)+(-0.5,0.15)$) rectangle ($(host.south east)+(0.5,-0.15)$);  
 \end{tikzpicture}

A given entity must be annotated with one label. The annotator must therefore choose the most relevant category according to the semantics of the message. We can therefore find in the dataset an entity annotated with different labels. For instance, Facebook can be categorized as a \textit{media} (``notre page Facebook") as well as an organization (``Facebook acquires acquiert Nascent Objects").

Event-named entities must include the type of the event. For example, \textit{colloque} (\textit{colloquium}) must be annotated in ``le \textbf{colloque du Réveil français} est rejoint par".

Abbreviations must be annotated. For example, \textit{LMP} is the abbreviation of ``Le Meilleur Patissier" which is a \textit{tvshow}.

As shown in Figure 1, the training and the test set have a similar distribution in terms of named entity types. The training set contains 2,902 entities among 1,656 unique entities (i.e. 57,1\%). The test set contains 3,660 entities among 2,264 unique entities (i.e. 61,8\%). Only 15,7\% of named entities are in both datasets (i.e. 307 named entities)\footnote{35\% in the case of CoNLL dataset.}. Finally we notice that less than 2\% of seen entities are ambiguous on the testset.

\section{Description of the Systems}
Overall, the results of 8 systems were submitted for evaluation. Among them, 7 submitted a paper discussing their implementation details. The participants proposed a variety of approaches principally using Deep Neural Networks (DNN) and Conditional Random Fields (CRF). In the rest of the section we provide a short overview for the approaches used by each system and discuss the achieved scores.

\textbf{Submission 1} \cite{Mourad17} The system relies on a recurrent neural network (RNN). More precisely, a bi-directional GRU network is used and a  CRF layer is adde on top of the network to improve  label prediction given information from the context of a word, that is the previous and next tags.

\textbf{Submission 2} \cite{Ngoc17} The system follows a state-of-the-art approach by using a CRF for to tag sentences with NER tags. The authors develop a set of features divided into six families (orthographic, morphosyntactic, lexical, syntactic, polysemic traits, and language-modeling traits).

\textbf{Submission 3}
\cite{Sileo17}, ranked first, employ CRF as a learning model. In the feature engineering process they use morphosyntactic features,  distributional ones as well as word clusters based on these learned representations.

\textbf{Submission 4} \cite{Ngoc17b}
The system also relies on  a CRF classifier operating on features extracted for each word of the tweet  such as  POS tags etc. In addition, they  employ an existing pattertn mining NER system (mXS) which is not trained for tweets. The addition of the system's results in improving the recall at the expense of precision.

\textbf{Submission 5} \cite{Nicole17}
The authors propose a bidirectional LSTM neural network architecture
embedding words, capitalization features and character embeddings learned
with convolutional neural networks. This basic model is extended
through a transfer learning approach in order to leverage English tweets
and thus overcome data sparsity issues.

\textbf{Submission 6}\cite{Bechet17}
The approach proposed here used adaptations for tailoring a generic NER system in the context of tweets. Specifically, the system is based on CRF and relies on features provided by context, POS tags, and lexicon. Training has been done using CAP data but also ESTER2 and DECODA available data. Among possible combinations, the best one used CAP data only and largely relied on a priori data.

\textbf{Submission 7}
Lastly, \cite{Seffih17} uses a rule based system which performs several linguistic analysis like morphological and syntactic as well as the extraction of relations. The dictionaries used by the system was augmented with new entities from the Web. Finally, linguistics rules were applied in order to tag the detected entities.

\section{Results}
Table \ref{tbl:res} presents the ranking of the systems with respect
to their F1-score as well as the precision and recall scores.

The approach proposed by \cite{Sileo17} topped the ranking showing how a standard CRF approach can benefit from high quality features. On the other
hand, the second best approach does not require heavy feature engineering as it relies on DNNs \cite{Mourad17}.

We also observe that the majority of the systems obtained good scores
in terms of F1-score while having important differences in precision and recall. For example, the Lattice team achieved the highest precision score.

\label{sec:results}
\begin{table}
    \begin{tabular}{p{3cm}rrr}\toprule
	System name & Precision & Recall & F1-score \\ \midrule
	Synapse Dev. & 73.65 & 49.06 & \textbf{58.59} \\
	Agadir & 58.95 & 46.83 & 52.19 \\
	TanDam & 60.67 & 45.48 & 51.99 \\
	NER Quebec & 67.65 & 41.26 & 51.26 \\
	Swiss Chocolate & 56.42 & 44.97 & 50.05\\
	AMU-LIF & 53.59 & 40.63 & 46.21 \\
	Lattice & 78.76 & 31.95 & 45.46 \\
	Geolsemantics & 19.66 & 23.18 & 21.28 \\ \bottomrule 
    \end{tabular}
    \caption{The scores for the evaluation measures used in the challenge for each of the participating systems. The official measure of the challenge was the micro-averaged $F_1$ measure. The best performance is shown in bold.}
    \label{tbl:res}
\end{table}

\section{Conclusion}
\label{sec:conclusion}
In this paper we presented the challenge on French Twitter Named Entity Recognition. A large corpus of around 6,000 tweets were manyally annotated for the purposes of training and evaluation. To the best of our knowledge this is the first corpus in French for NER in short and noisy texts. A total of 8 teams participated in the competition, employing a variety of state-of-the-art approaches. The evaluation of the systems helped us to reveal the strong points and the weaknesses of these approaches and to suggest potential future directions. 
\bibliography{cap2017}

\end{document}